%% file: main.tex
\documentclass[conference]{IEEEtran}
\IEEEoverridecommandlockouts
\usepackage{cite}
\usepackage{amsmath,amssymb,amsfonts}
\usepackage{algorithmic}
\usepackage{graphicx}
\usepackage{textcomp}
\usepackage{xcolor}

\usepackage{verbatim}
\usepackage{amsmath}
\usepackage{amssymb}
\usepackage{booktabs}
\usepackage{hyperref}

\hypersetup{
    colorlinks=true,
    linkcolor=black,
    citecolor=black, % citation
    pdfpagemode=FullScreen,
    pdftitle={VOD-RAISE},
    urlcolor=black,
    }
    
\newcommand{\metatable}{\text{meta-table}}

\makeatletter
\newcommand{\linebreakand}{%
  \end{@IEEEauthorhalign}
  \hfill\mbox{}\par
  \mbox{}\hfill\begin{@IEEEauthorhalign}
}
\makeatother

\def\BibTeX{{\rm B\kern-.05em{\sc i\kern-.025em b}\kern-.08em
    T\kern-.1667em\lower.7ex\hbox{E}\kern-.125emX}}
\begin{document}
\title{A 2-Stage Model for Vehicle Class and Orientation Detection with Photo-Realistic Image Generation\\
\thanks{\\\IEEEauthorrefmark{1}Co-leading authors.\\\IEEEauthorrefmark{2}Corresponding author. }
}

%\author{\IEEEauthorblockN{1\textsuperscript{st} Youngmin Kim}
%\and
%\IEEEauthorblockN{2\textsuperscript{nd} Donghwa Kang}
%\and
%\IEEEauthorblockN{3\textsuperscript{rd} Hyeongboo Baek}
%\IEEEauthorblockA{\textit{Dept. of Computer Science and Engineering}\\
%\textit{Incheon National University(INU)}\\
%Incheon, Republic of Korea \\
%winston121497@gmail.com}
%\IEEEauthorblockA{\textit{Dept. of Computer Science and Engineering} \\
%\textit{Incheon National University(INU)}\\
%Incheon, Republic of Korea \\
%anima0729@inu.ac.kr}
%\linebreakand 
%\IEEEauthorblockN{3\textsuperscript{rd} Hyeongboo Baek}
%\IEEEauthorblockA{\textit{Dept. of Computer Science and Engineering} \\
%\textit{Incheon National University(INU)}\\
%Incheon, Republic of Korea \\
%hbbaek@inu.ac.kr}
%}

\author{
    \IEEEauthorblockN{Youngmin Kim\IEEEauthorrefmark{1}, Donghwa Kang\IEEEauthorrefmark{1}, Hyeongboo Baek\IEEEauthorrefmark{2}}
    \IEEEauthorblockA{ Dept. of Computer Science and Engineering Incheon National University(INU) Incheon, Republic of Korea
    \\\{winston1214, anima0729, hbbaek\}@inu.ac.kr}
}
\IEEEoverridecommandlockouts
\IEEEpubid{\makebox[\columnwidth]{978-1-6654-8045-1/22/\$31.00~\copyright2022 IEEE\hfill} \hspace{\columnsep}\makebox[\columnwidth]{ }}
\maketitle
\IEEEpubidadjcol
\input{00abstract}

\input{01introduction}
\begin{figure}[t!]
    \centering
    \includegraphics[width=1\columnwidth]{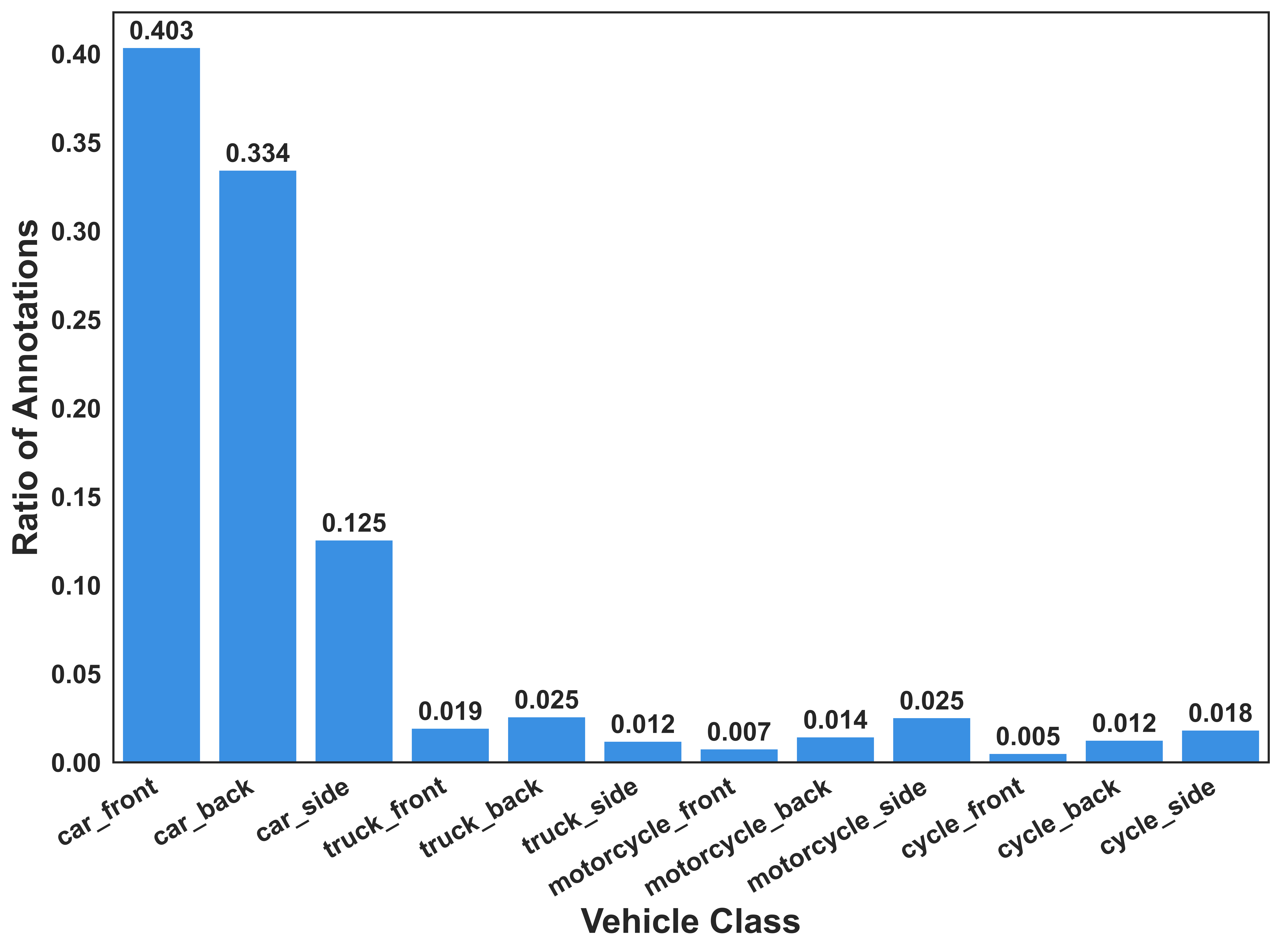}
    \caption{Number of annotations by class}
    \label{fig-label1}
\end{figure}
\begin{figure*}[t!]
    \centering
    \includegraphics[width=1\linewidth]{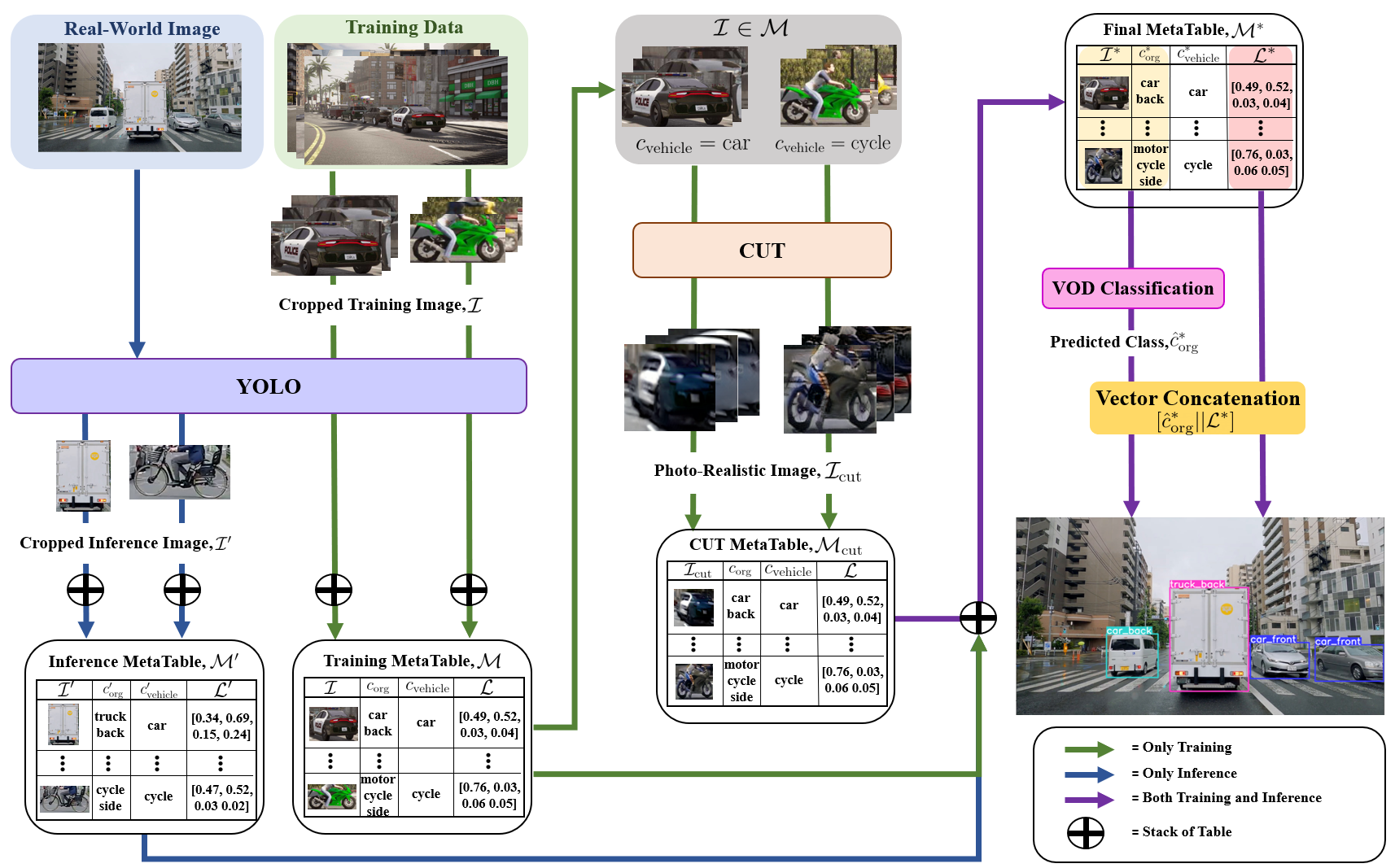}
    \caption{Overview of our proposed VOD approach.}
    \label{fig-label2}
\end{figure*}
\input{02dataset}

\input{03method}

\input{04results}

\input{06conclusion}

\section*{Acknowledgement}
This work was supported by the National Research Foundation of Korea (NRF) grant funded by the Korea government (MSIT) (NRF-2022R1A4A3018824).

\bibliographystyle{ieeetr}
\bibliography{ref}
\end{document}

%% file: 00abstract.tex
\begin{abstract}
\begin{comment}
우리는 합성 데이터로 모델을 학습시켜 차량의 종류와 방향을 탐지하는 것을 목적으로 한다.
그러나 훈련 데이터가 클래스의 분포가 불균형하고, 합성이미지를 훈련시킨 모델은 실제 환경의 이미지에서 예측이 어렵다.
우리는 이러한 문제를 해결하기 위해, photo-realistic 이미지 생성을 통한 2-stage 탐지 모델을 제안한다.
우리의 모델은 차량 종류와 방향을 탐지하기 위해 주로 4가지의 step을 거친다.
(1) 이미지에 있는 객체의 이미지,클래스,위치 정보를 담은 테이블을 구축하고, 
(2) 합성 이미지들을 실제 이미지 스타일로 변환한 후, 이를 메타테이블에 합친다.
(3) 메타테이블에 있는 이미지들을 이용하여 차량 클래스 및 방향을 예측하고,
(4) 미리 추출한 지역 정보들과 예측한 클래스를 합쳐 차량 클래스 및 방향 탐지를 이뤄낸다.
우리는 우리의 접근 방법으로 IEEE BigData Challenge 2022 Vehicle class and orientation detection에서 4등을 달성하였다.
\end{comment}
We aim to detect the class and orientation of a vehicle by training a model with synthetic data.
However, the distribution of the classes in the training data is imbalanced, and the model trained on the synthetic image is difficult to predict in real-world images.
We propose a two-stage detection model with photo-realistic image generation to tackle this issue.
Our model mainly takes four steps to detect the class and orientation of the vehicle.
(1) It builds a table containing the image, class, and location information of objects in the image, (2) transforms the synthetic images into real-world images style, and merges them into the meta table.
(3) Classify vehicle class and orientation using images from the meta-table. 
(4) Finally, the vehicle class and orientation are detected by combining the pre-extracted location information and the predicted classes.
We achieved 4\textsuperscript{th} place in IEEE BigData Challenge 2022 Vehicle class and Orientation Detection (VOD) with our approach.
Our code and project material will be available at \url{https://github.com/inu-RAISE/VOD\_Challenge}.
\end{abstract}
\begin{IEEEkeywords}
Object Detection, Classification, Image-to-Image Translation, GAN
\end{IEEEkeywords}

%% file: 01introduction.tex
\section{Introduction}
\begin{comment}
Computer Vision 분야에서 각광받고 있는 Transformer\cite{transformer}나 CNN\cite{CNN} 기반의 모델을 사용하여 훈련하기 위해선 많은 양의 데이터가 필요하다.
또한, 이러한 모델을 이용하여 real world에 적용하기 위해선 실제 환경의 데이터를 수집하고 구성해야한다.
그러나 이 작업은 많은 시간과 비용이 소요되는 문제점이 있다.
우리는 이러한 문제점을 해결하기 위해 비교적 비용이 덜 소요되는 가상의 합성 데이터를 활용하여 학습을 한 뒤, 실제 이미지에서 예측을 보다 잘 할 수 있는 방법을 제안한다.

우리의 제안방법은 다음과 같은 과정을 따른다. 
먼저, 이미지의 객체 정보를 담은 테이블을 구축하고, 객체의 위치 정보를 추출한다.
이후, 합성된 객체 이미지들을 실제 이미지처럼 스타일을 변환한다.
변환된 photo-realistic 이미지와 합성 이미지들을 이용하여 차량의 클래스와 방향을 분류하고, 위치 정보를 결합하여 실제 이미지에서의 탐지를 이뤄낸다.
이러한 방법으로 우리는 오직 합성 데이터로만 훈련을 한 기존의 객체 탐지 모델보다 $8\%p$ 높은 정확도를 얻었다.

본 논문의 기여는 다음과 같다.
- 우리는 이미지에서 각 객체에 대한 정보를 담은 meta-table을 구축하여 차량의 클래스와 방향을 탐지하는 2-stage 모델을 제안한다.
- 우리는 photo-realistic 이미지 생성으로 실제 환경에서 예측을 보다 잘 할 수 있는 방법을 제안한다.

본 논문에선 차량의 클래스와 방향을 탐재하는 우리의 제안한 접근법을 소개하고, 논문의 나머지 부분은 다음과 같이 구성된다.
섹션 2에선, 우리는 합성 데이터셋과 객체 탐지에 대한 관련 연구들을 소개하고, 섹션 3에선 우리의 상세한 접근법을 소개한다.
섹션 4와 5에선 우리의 실험 환경과 실험 결과를 공유하고, 마지막으로 section 6에선 본 논문의 결론과 한계점을 논의한다.

%이러한 배경 하에, VOD challenge는 synthetic dataset에서 훈련된 모델을 사용하여 실제 이미지에 대한 객체 탐지 모델을 개선할 수 있는 방향을 다양한 기술을 탐색하는 대회이다.
%우리는 이 challenge에 참여하였고 4등을 하였다.
%본 논문에선, 우리의 접근 방법을 보여준다.
% 새로운 말들

\end{comment}
A large amount of data is required to train using a Transformer\cite{transformer} or Convolution Neural Networks (CNN)\cite{CNN} based model, which is in the spotlight in the field of computer vision.
In addition, to apply this model to the real world using this model, it is necessary to collect and construct data from the real environment.
However, this work has a challenge because it takes a lot of time and cost.
To solve this challenge, we study using virtual synthetic data, which is relatively inexpensive, and then study to improve the model that can operate in a real-world driving environment.

Our proposed approach follows:
First, a table containing the object information of the image is built, and the location information of the object is extracted.
After that, the styles of the synthesized object images are transformed into photo-realistic images.
Using the transformed photo-realistic images and synthetic images, the vehicle class and orientation are classified, and location information is combined to achieve detection in real-world images.
In this way, we achieved an accuracy that is about $8\%p$ higher than that of the existing object detection model trained only on synthetic data.

This paper is summarized as follows:
\begin{itemize}
    \item We propose a two-stage model that detects the class and orientation of a vehicle by building a meta-table containing information about each object in images.
    \item We propose a method for better prediction in the real world with photo-realistic image generation.
\end{itemize}

In this paper, we introduce our proposed approach to detect vehicle class and orientation, and the rest of the paper is organized as follows:
In Section 2, we survey synthetic datasets and related studies on object detection. In Section 3, we introduce our detailed approach.
We share our experimental environment and experimental results in Sections 4 and 5, respectively. 
Finally, we conclude our paper in Section 6 by discussing our findings and the limitations of our paper.

%% file: 02dataset.tex
\section{Related Works}
\subsection{Synthetic Dataset}
\begin{comment}
%이 이미지 데이터셋은 IEEE BigData Cup Challenge 2022에서 차량의 종류와 방향을 탐지하기 위해 제공한 것이다.
Object Detection 분야에서 실세계의 이미지와 객체에 맞는 label을 수집하는 것은 비용이 많이 든다.
이 때문에 비교적 비용이 덜 드는 데이터셋을 수집하고 object detection을 수행하기 위해서 비교적 비용이 덜 드는 가상의 합성 데이터셋을 이용한다.

Kumar et al은 교통 흐름을 파악하기 위해 데이터셋을 수집하였고, 이는 차량의 종류와 방향에 대한 class와 위치 정보를 포함한다.
이 데이터 세트에는 1920 x 1080 해상도에서 교육을 위한 총 63,066개의 합성 이미지가 있고 테스트 데이터에는 총 3000개의 이미지에 대해 각각 1920 x 1080 및 1280 x 720 해상도의 1500개 실제 이미지의 두 가지 버전이 있다.
데이터셋의 class는 총 12개로 4개의 vehicle의 종류(car,truck,motorcycle, cycle)와 3가지의 방향(back,front,side)으로 구성되어 있다.
각 class 별 데이터의 수는 Fig 1에 나타나있다.

FCAV dataset은 가상의 데이터로 비디오 게임 기반 시뮬레이션 엔진의 이미지로 GTA5를 기반으로 만들어졌다.
이 데이터셋은 4가지의 날씨(sun, fog, rain and haze)와 4가지의 시간대(day, night, morning and dusk)에서 운전하는 상황을 가정하여 수집되어졌기 때문에 다양한 상황의 이미지를 포함한다.
FCAV의 훈련 데이터셋은 총 205,879개이고, 3개의 class (car,motorbike,person)로 구성되어있다.
각 class 별로 사람은 58 objects, 차 class는 1,125,624 objects, motorbike는 15,902 objects들이 있다.
이 데이터셋에서의 car는 우리가 truck이라고 정의한 것도 포함이 되어있다.
본 논문에선 FCAV 데이터셋의 훈련 데이터셋만 사용하였고, person class는 제외하였다.
\end{comment}
Collecting labels that fit images and objects in the real world is an expensive and challenging process in object detection using deep learning.
For this reason, a relatively inexpensive synthetic dataset is used to collect a relatively inexpensive dataset and perform object detection.
Kumar \textit{et al}.\cite{dataset1,bigcup,bigcup2} collected a dataset that includes vehicle class and orientation and location information to identify traffic flow.
This image dataset has a total of 63,066 synthetic images for training at 1920 x 1080 resolution.
Also, a total of 3,000 real-world images were provided for testing at 1920 x 1080 and 1280 x 720 resolutions, a number of 1,500, respectively.
There are 12 classes in the dataset, consisting of four types of vehicles (car, truck, motorcycle, cycle) and three orientations (back, front, side).
The number of annotations for each class is shown in Fig. \ref{fig-label1}.

The FCAV dataset\cite{FCAV} is an image of a video game-based simulation engine as virtual data, and was created based on GTA \uppercase\expandafter{\romannumeral5}\cite{GTA}.
This dataset contains images of various situations because it was collected assuming driving in 4 weather conditions (sun, fog, rain, haze) and 4-time ranges (day, night, morning, dusk).
 There are a total of 205,879 training datasets for FCAV, consisting of three classes (person, car, and motorbike).
For each class, there are 58 objects for a person, 1,125,624 objects for a car class, and 15,902 objects for a motorbike.
We use only the training dataset, and the person class is excluded.

\subsection{Object Detection}
\begin{comment}
객체 탐지 작업은 이미지에서 특정 클래스의 개체 인스턴스를 감지하여 클래스 분류와 객체 지역화를 동시에 수행하는 것이다.
객체 탐지 작업에는 크게 두가지로 나뉜다.
클래스 분류와 객체 지역화를 한 번에 수행하는 1-stage 모델과, 객체 지역화를 한 후 클래스 분류를 진행하는 2-stage 모델이 있다.
1-stage 모델은 추론 속도가 비교적 빠르고 대표적인 모델은 SSD와 YOLO가 있다.
반면에 2-stage 모델은 정확도가 비교적 높고 대표적인 모델은 RCNN 계열이 있다.

우리는 COCO 데이터셋에서 높은 정확도를 가지고, 빠른 추론시간을 가지는 YOLO의 최신 버전인 YOLOv5를 사용하였다.
\end{comment}
The object detection task is to detect object instances of a specific class in images to perform both class classification and object localization.
The methods for object detection can be categorized into two main types: There is a single-stage model that performs class classification and object localization simultaneously, and a two-stage model proceeds with class classification after object localization.
The single-stage model has a relatively fast inference speed, and representative models include SSD\cite{ssd} and YOLO\cite{yolo}.
On the other hand, the two-stage model has relatively high performance, and the representative model is the RCNN series\cite{rcnn,fastrcnn,faster}.

We used YOLOv5\footnote{https://github.com/ultralytics/yolov5}, the latest version of YOLO, which has high accuracy and fast inference time on the COCO dataset\cite{coco}.

%% file: 03method.tex
\section{Proposed Approach}
\begin{comment}
우리는 훈련 데이터셋의 클래스가 불균형하게 이뤄져 학습의 편향성이 발생하는 것을 방지하기 위해, 지역 정보와 vehicle의 종류를 분류하는 문제를 각각 나누어 해결하는 방법을 사용하였다.
또한, 실제 환경에서 촬영한 이미지에서 예측을 잘 할 수 있도록 합성 이미지를 실제같은 이미지로 변환하였다.
이를 위해, 우리는 훈련 데이터의 label을 이용하여 객체별로 이미지를 잘랐고, n개의 객체에 대한 정보를 가지는 Metatable M을 정의한다.
그것들은 객체의 이미지, 클래스, 위치 정보로 이뤄지고 이는 $\mathcal{I}, \mathcal{C},\mathcal{L} \subset \mathcal{M}$를 따른다.

$\mathcal{I} = \{\mathcal{I}_1 , \cdots \mathcal{I}_n \}$는 객체의 이미지와 그에 대한 label, 그리고 위치 정보로 구성된다. % 식으로 나타내고, where 써서 notation 정의
\metatable을 이용한 우리의 접근 방식은 그림 1에 나와있고, 다음의 4가지 방식으로 요약이 된다.
1) 훈련 이미지에서 추출한 객체의 이미지, 클래스, 위치 정보를 메타 테이블에 쌓는다.
2) 합성 이미지를 사실적인 이미지로 변환하고, 새로운 메타 테이블을 만든다.
3) 새로운 메타테이블과 훈련데이터의 메타 테이블을 결합한 메타테이블에서 이미지와 클래스로 이미지 분류 모델로 차량의 종류와 방향을 분류하고,
4) 예측된 클래스와 메타테이블의 위치정보를 결합하여 차량의 종류와 방향 탐지를 이뤄낸다.
\end{comment}
To avoid learning bias due to imbalanced classes in the training data set, we used a method to solve the problem by separating it into regional proposals and vehicle class classification.
Also, we transform the synthetic image into a photo-realistic image to make better predictions from real-world images.
For this, we crop the image by object using the label of the training data and define training \text{\metatable}, $\mathcal{M} = \{\mathcal{M}_1, \cdots, \mathcal{M}_n\}$, with information about the number of $n$ objects.
It is built with the object's images dictionary, $\mathcal{I} = [\mathcal{I}_1 , \cdots, \mathcal{I}_n]$, classes dictionary, $\mathcal{C} = [\mathcal{C}_1 , \cdots, \mathcal{C}_n]$, and location dictionary, $\mathcal{L} = [\mathcal{L}_1 , \cdots, \mathcal{L}_n]$, according to $\mathcal{I}, \mathcal{C},\mathcal{L} \subset \mathcal{M}$.

Our approach using \text{\metatable} is illustrated in Figure 1 and summarized in the following four steps.
1) It stacks the image, class, and location information of the object extracted from the training image in the \text{\metatable}.
2) Transform synthetic images to photo-realistic images and creates a new \text{\metatable}, $\mathcal{M}_\text{cut}$.
3) The image classification model classifies the class and orientation of the vehicle by the image and class in the final meta-table $\mathcal{M}^{*}$, which combines $\mathcal{M}_\text{cut}$ and the meta-table $\mathcal{M}$ of the training data.
4) Finally, the vehicle class and orientation are detected by combining the predicted class and the location information of the $\mathcal{M}^{*}$.
\begin{figure}[t!]
    \centering
    \includegraphics[width=1\linewidth]{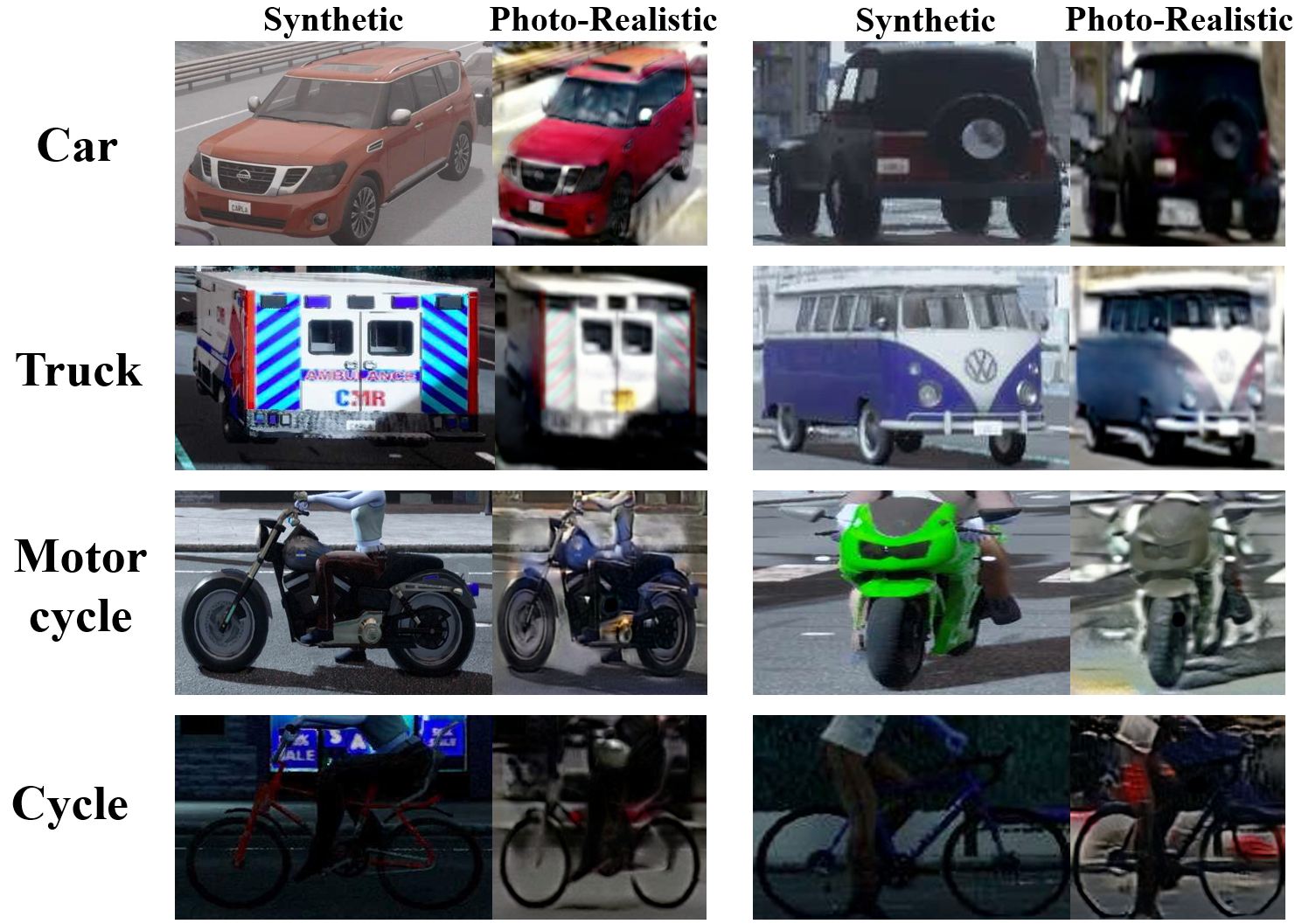}
    \caption{Examples of transform result from synthetic image to photo-realistic image}
    \label{fig-label3}
\end{figure}
\subsection{Meta-table Representation} % working title (YOLOv5) 부분
\subsubsection{Training} % 구성 대신 표현이라는 단어로 변환할 것을 고려해볼것
\begin{comment}
주어진 데이터셋의 라벨 정보로 객체의 이미지 I와 위치 R, 그리고 클래스 C를 구성한다.
먼저 주어진 정규화된 객체의 크기 정보($(x,y,w,h) \in \mathcal{L}$)로 훈련 이미지에서 객체만 자른 이미지들을 I에 쌓는다.
그리고 vehicle의 종류와 방향이 포함된 12개의 class $c_{org}$와 차 또는 motorbike로 정의된 class $c_{vehicle}$로 class 집합을 나눈 후, C에 쌓고 이는 $c_i = [c_{org},c_{vehicle}] \in \mathcal{C}$를 따른다. 

In summary, $i$-th $m$ is defined as:
\begin{equation}
    m_i = [\mathcal{I}_i , x^{i}_\text{{min}},y^{i}_\text{{min}},x^{i}_\text{{max}},y^{i}_\text{{max}},c^{i}_\text{{org}},c^{i}_\text{{vehicle}}]
\end{equation}
FCAV 데이터셋도 같은 방법으로 dictionary를 구성한 후, M에 쌓는다.
이렇게 구성한 $\mathcal{M}$을 이용하여 Object Detection 모델인 YOLOv5를 훈련시킨다.
%% 여기에 들어가야될 말
우리는 이 때, 전체 클래스 정보가 담겨있는 $c_{org}$가 아닌 binary class 정보인 $c_{vehicle}$을 이용한다.
Fig 1을 보면 훈련 데이터셋은 불균형한 분포를 가지기 때문에 우리는 이진 클래스로 객체를 탐지하는 것이 아닌, 각각 하나의 클래스를 예측하는 방법을 이용한다.
즉, car와 motorcycle에 대해 객체 탐지를 각각 수행한다.
이를 통해 비교적 객체의 수가 적은 cycle과 motorcycle에 대한 위치 정보를 보다 정확히 파악할 수 있고, 이는 데이터의 불균형을 해소하는 방법이다.
\end{comment}
It builds the training image dictionary, $\mathcal{I} = [\mathcal{I}_1 , \cdots , \mathcal{I}_n]$, location dictionary of training data, $\mathcal{L} = [\mathcal{L}_1 , \cdots , \mathcal{L}_n]$, and class dictionary, $C = [\mathcal{C}_1 , \cdots , \mathcal{C}_n]$, of the object with the given label information of the training dataset.
First, object images obtained by cropping only the object from the training datasets with the given normalized object location information ($(x,y,w,h) \in \mathcal{L}$) stacked on $\mathcal{I}$.
Then, after dividing the class dictionary into 12 classes, $c_{org}$, including the class and direction of the vehicle and the class, $c_{vehicle}$, defined as the car or motorbike, stack them in $C$.
It follows $\mathcal{C}_{i} = [c^{i}_\text{{org}},c^{i}_\text{{vehicle}}] \in \mathcal{C}$, where $i$ means $i$-th.
In summary, $i$-th $\mathcal{M}$ follows:
\begin{equation}
    \mathcal{M}_i = [\mathcal{I}_i , x_{i},y_{i},w_{i},h_{i},c^{i}_\text{{org}},c^{i}_\text{{vehicle}}]
\end{equation}
The FCAV dataset\cite{FCAV} is also stacked in $M$ after constructing a dictionary in the same way.
Using the $\mathcal{M}$ configured in this way, YOLOv5, an object detection model, is trained.
At this time, we use $c_\text{{vehicle}}$, which is binary class information, not $c_\text{{org}}$, which contains all class information.
As shown in Fig.\ref{fig-label1}, since the training dataset has an imbalanced distribution, we do not predict objects as binary classes but instead, predict one class each.
i.e., object detection is performed about cars and motorcycles, respectively.
Through this, it is possible to more accurately predict the location information about cycles and motorcycles with a relatively small number of objects, which is a method of resolving the data imbalance.

\subsubsection{Inference}
\begin{comment}
훈련 과정과 달리 Inference 과정은 객체에 대한 제공된 위치 정보와 class 정보가 없기 때문에 바로 \text{\metatable}을 구성할 수 없다.
따라서 자동차 클래스와 오토바이 클래스를 구분하여 사전 훈련된 YOLOv5 모델을 사용하여 두 클래스 각각에 대한 객체를 예측한다. 
새로운 R과 C는 YOLOv5로 예측된 클래스와 로컬 정보를 기반으로 표현된다.
이 때, C'는 훈련 과정과 달리 이진 class인 $c_{vehicle}$로만 표현된다.
그리고 지역 정보 사전인 R'을 이용하여 이미지에서 객체별로 자른 후, 자른 이미지로 사전 I'를 만든다.
정리하자면, 추론 과정에서의 meta-table은 이를 따른다.
\end{comment}
Unlike the training process, the inference process cannot directly build \text{\metatable} because there is no provided location information and class information about objects.
Therefore, by distinguishing the car and motorcycle classes, the pretrained YOLOv5 model predicts the class and location of objects for each of the two classes. 
The new inference dictionary $\mathcal{L}'$ and $\mathcal{C}'$ are represented based on class and location information predicted by YOLOv5.
At this time, $\mathcal{C}'$ is represented only by binary class $c_\text{vehicle}'$, unlike the training process.
Then, the image is cropped by objects using $\mathcal{L}'$, a location information dictionary, and a dictionary $\mathcal{I}'$ is made from the cropped images.
In summary, the meta-table in the inference process, $\mathcal{M}'$ follows: % inference object 개수 정의해야되지 않을까
\begin{equation}
    \mathcal{M}' = [\mathcal{I}' , \mathcal{L}',c_{\text{vehicle}}']    
\end{equation}
\subsection{Synthetic Image to Photo-Realistic Image} % synthetic vs synthesis 통일하기
\begin{comment}
실제 환경과 인위적으로 만들어진 합성 이미지는 분명한 이미지간의 차이가 존재한다.
이 때문에 합성 이미지로 모델을 훈련시키면 real-world 이미지에 적용이 비교적 잘 되지 않는다.
따라서 우리는 가상의 이미지를 사실적인 이미지로 변환시키는 이미지 대 이미지 번역을 통해 분류 모델의 학습에 도움을 준다.

Pix2Pix와 같은 이미지 대 이미지 번역은 정렬된 이미지 쌍의 훈련 세트를 사용하여 입력 이미지와 출력 이미지 간의 매핑을 학습한다.
그러나 이미지 쌍의 훈련 세트를 수집하는 것은 매우 어려운 일이고, 비용이 많이 든다.
그래서 우리는 쌍을 이루지 않은 이미지 훈련 세트를 사용할 수 있는 이미지 대 이미지 번역 모델을 사용하였다.
쌍을 이루지 않는 이미지 대 이미지 번역 모델 중 우리는 self-supervised representation learning 분야에서 state of the arts를 달성한 CUT을 사용하였다.

우리는 $\mathcal{M}$에 있는 $\mathcal{I}$를 CUT을 이용하여 사실적인 이미지로 변환하였다.
이에 대한 몇몇 예시들은 Fig 3에 나타난다.
사실적인 이미지로 변환한 이미지들로 새로운 이미지 dictionary I를 구축하고, 각 이미지 즉, I와 I_cut이 연결되는 M의 클래스와 지역 정보들로 M_cut를 표현한다.
실제와 같은 스타일로 변환한 이미지의 정보를 담은 metatable M_cut은 다음을 따른다.
\end{comment}
There is a clear difference between real-world data and artificially created synthetic data.
For this reason, training a model with synthetic images does not apply well to real-world images.
Therefore, we help train the classification model through image-to-image translation that transforms synthetic images into a photo-realistic images.

Image-to-image translation model, such as Pix2Pix\cite{pix2pix}, uses a training set of aligned image pairs to learn mappings between input and output images.
However, collecting a training set of image pairs is very difficult and costly.
So we used an image-to-image translation model that can be used even on unpaired image training sets.
Among the unpaired image-to-image translation models, we used CUT\cite{cut}, which achieved a state of the arts (SOTA) in self-supervised representation learning.
We transformed the $\mathcal{I}$ in $\mathcal{M}$ into photo-realistic images using CUT.
Examples of the results are shown in Fig. \ref{fig-label3}.
A new image dictionary $\mathcal{I}_\text{cut}$ is built as images transformed into photo-realistic images, and $\mathcal{M}_\text{cut}=\{\mathcal{M}^{1}_\text{cut} , \cdots , \mathcal{M}^{n}_\text{cut}\}$ is represented as class, $\mathcal{C}$, and location information, $\mathcal{L}$ of $\mathcal{M}$ in which each image, i.e. $\mathcal{I}_\text{cut}$ and $\mathcal{I}$ are matched.
The \text{\metatable} $\mathcal{M}_\text{cut}$ that contains the information of the image converted to the style as real-world follows:
\begin{equation}
    \mathcal{M}^{i}_\text{cut} = [\mathcal{I}^{i}_\text{cut},\mathcal{C}^{i},\mathcal{L}^{i}]
\end{equation}

\subsection{Vehicle Class and Orientation Detection}
\begin{comment}
% notation을 사용하여 수정할 수 있도록 만들자
우리는 metatable을 이용하여 차량의 종류와 방향을 탐지한다.
최종적인 metatable M*는 훈련 시에는 M과 M_cut을 합치고, 추론 시에는 M_cut과 M'을 합쳐서 표현한다.
우리는 차량의 클래스와 방향에 대한 분류를 위해 이미지 분류 분야에서 높은 정확도를 가지는 EfficientNet을 사용한다.
EfficientNet은 다양한 데이터셋에서 높은 정확도를 유지하는 안정적인 이미지 분류 모델로 많은 SOTA 모델의 근간이 되는 모델이다.
따라서 EfficientNet으로 12개의 class에 대해 예측을 하고, 예측된 class, c*을 지역 정보 R*와 결합시켜 차량의 클래스와 방향에 대한 탐지를 이뤄낸다.
EfficientNet는 파라미터 개수에 따라서 b0부터 b7으로 나눠지는데 숫자가 커질수록 모델의 파라미터 수가 많아진다.
우리는 연산 속도가 느린 반면 정확도가 높은 b7을 선택하여 분류 정확도를 높이는데 초점을 맞췄다.
또한, 과적합 방지와 정확도를 향상 시키기 위해 5번의 앙상블 과정을 거쳐서 최종적인 분류 architecture를 설립하였다.
\end{comment}
We use \text{\metatable} to detect the class and orientation of the vehicle.
The final \text{\metatable}, $\mathcal{M}^{*} = \{\mathcal{I}^{*},c^{*}_\text{org},c^{*}_\text{vehicle}, \mathcal{L}^{*} \}$ is represented by combining $\mathcal{M}$ and $\mathcal{M}_\text{cut}$ during the training process.
Likewise in the inference process, $\mathcal{M}^{*}$ is expressed as a combination of $\mathcal{M}'$ and $\mathcal{M}_\text{cut}$.
We use EfficientNet\cite{eff} with high accuracy in the image classification task to classify vehicle class and orientation.
This model is a stable image classification model that maintains high accuracy in various datasets and is the base of many states the of art models.
Therefore, EfficientNet predicts 12 classes after that and combines the predicted class, $c^{*}_{org}$, with location information, $\mathcal{L}^{*}$, to detect the class and orientation of the vehicle.
EfficientNet is divided into b0 to b7 according to the number of parameters.
As the number increases, the number of parameters in the model increases.
We focused on improving classification accuracy by choosing b7, which has high accuracy but a slow processing speed.
In addition, to prevent overfitting and improve accuracy, the final classification architecture was designed through an ensemble process 5 times.

%% file: 04results.tex
\section{Experiments Setup}
\begin{comment}
우리는 VOD를 수행하기 위해 YOLOv5, CUT, EfficientNet 모델을 사용하였고, 각 모델에 따른 하이퍼파라미터가 다양하다.
먼저 YOLOv5는 We fine-tuned the YOLOv5x6 model which is a pretrained weight on COCO dataset.
And we set 이미지 사이즈를 640 x 482로 설정하였고, 배치 사이즈는 16, epoch은 20으로 설정하였다.
다음으로 CUT은 예측된 객체의 크기 또는 주어진 객체의 크기가 다양하므로 이미지의 크기는 256 x 256으로 epoch은 100으로 설정하였고, 사전 훈련된 모델을 사용하지 않고 처음부터 훈련시켰다.
마지막으로 ImageNet으로 pretrained 된 모델을 사용하였고, 파라미터 수가 많아 오버피팅이 발생하는 것을 방지하기 위해 이미지를 회전과 노이즈를 주면서 증강시켰다.
그리고 이미지 사이즈는 384 x 384로 설정하여 훈련과 추론을 진행한다.
모든 실험은 A6000 GPU가 2개 있는 환경에서 수행하였다.
\end{comment}
\begin{figure}[t!]
    \centering
    \includegraphics[width=1\columnwidth]{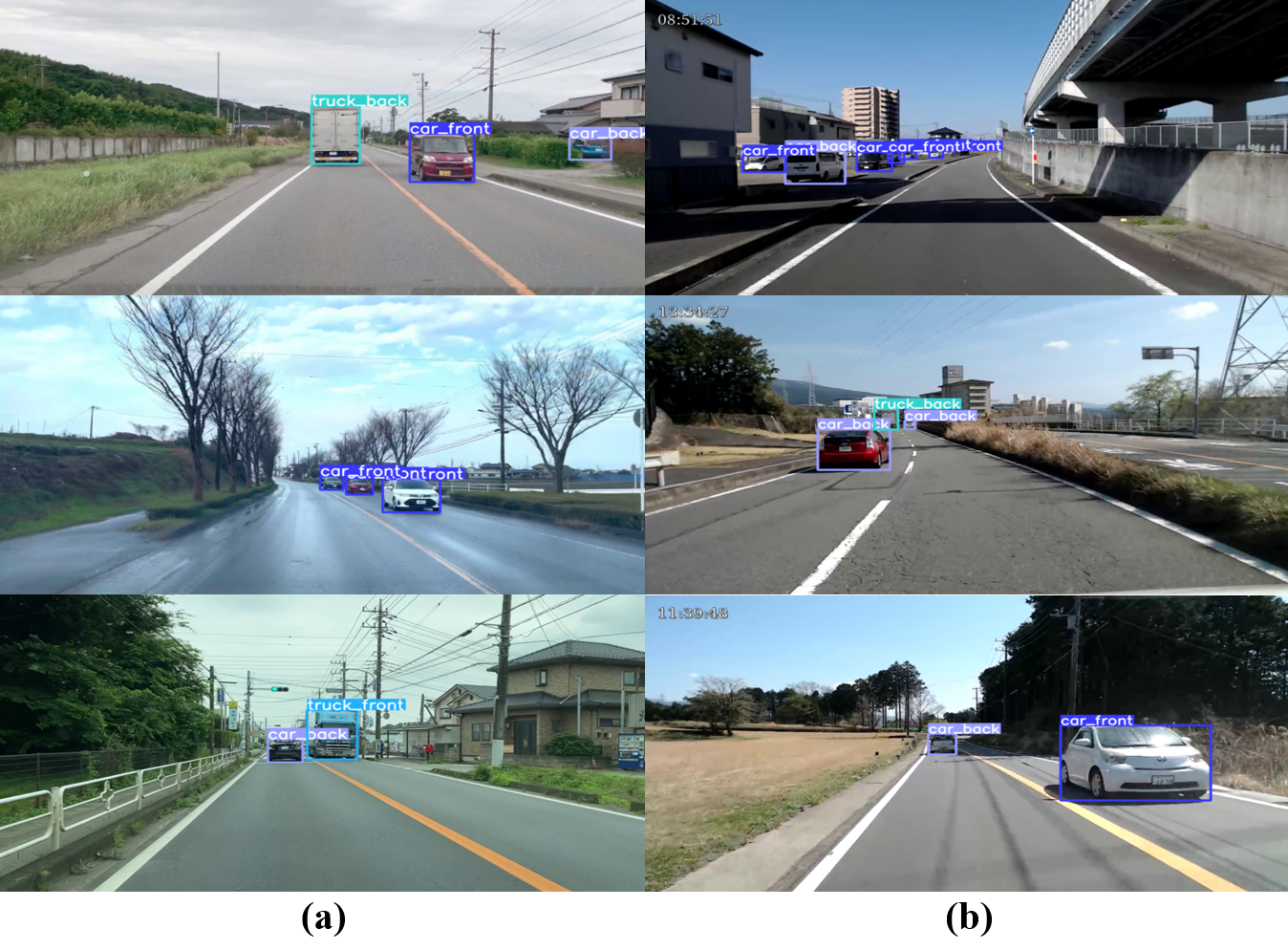}
    \caption{Examples of predicted vehicle class and direction detection result. (a) is the result images for the first test set and (b) is the result images for the second test set.}
    \label{fig-label4}
\end{figure}
We used YOLOv5, CUT, and EfficientNet models to perform VOD, and the hyperparameters for each model are varied.
First, YOLOv5 fine-tuned the YOLOv5x6 model, a pretrained weight on the COCO dataset.
Furthermore, we set the image size to 640 x 482, batch size to 16, and epoch to 20.
Next, since the size of the predicted object or a given object varies in the CUT, the image size is set to 256 x 256, the epoch is set to 100, and it is trained from scratch without using a pretrained model.
Finally, a model pretrained with ImageNet is used, and the image is augmented with rotation and noise to prevent overfitting due to many parameters.
The image size is set to 384 x 384 to proceed with training and inference.
All our experiments were performed in an environment with two A6000 GPUs.
\section{Results}
\begin{comment}
우리는 평가 지표를 VOD Challenge의 metric으로 사용하였고, 이는 다음을 따른다.
\begin{eqnarray}
    WmAP = \sum^{12}_{k=1} w^{k} \times AP ; where \sum^{12}_{k=1} = 1 \\
    score = 0.45 \times \underset{v_1}{WmAP} + 0.55 \times \underset{v_2}{WmAP}
\end{eqnarray}
이 평가지표는 각 class와 데이터셋에 따라 가중치를 다르게 주어 최종적인 점수를 도출한다.
Table 2에는 vanilla yolov5로 예측을 한 것과 우리의 방식을 비교한 결과가 나타나있다.
%그냥 YOLO 로만 했을 떄랑 비교
%우리 등수 비교 (전체 등수 쓰고)
Table 1에서 볼 수 있듯이,합성 데이터로만 훈련시킨 YOLOv5로 예측했을 때보다 우리의 접근 방식이 더욱 좋다는 것을 입증하였다.
YOLOv5는 합성 데이터로만 훈련시켰기 때문에 실제 환경 이미지에서 모델이 예측을 잘 할 수 없었고, 더불어 localization과 12개의 class를 한 번에 분류하는 것을 동시에 하였기 때문에 모델이 어려운 문제를 풀 수 밖에 없어서 예측을 못하였다.
반면에 우리의 접근 방법은 2-stage로 구성하여 localization과 classification을 별개로 수행하여 모델이 쉬운 문제를 풀 수 있도록 하였고, photo-realistic 이미지를 생성한 후 훈련하여 real-world image에서도 예측을 할 수 있도록 하였다.
우리의 접근 방법을 이용한 예측된 클래스와 경계상자의 예시는 Fig 4에 보여진다.
\end{comment}
We report the performance of our approach on the VOD challenge evaluation system, with metrics as follows:
\begin{eqnarray}
    WmAP = \sum^{12}_{k=1} w^{k} \times AP ; \; where, \sum^{12}_{k=1} = 1 \\
    score = 0.45 \times \underset{v_1}{WmAP} + 0.55 \times \underset{v_2}{WmAP}
\end{eqnarray}
This metric derives the final score by giving different weights for each class and data set.
In this Table \ref{table1}, we list the results of comparing the prediction made with vanilla YOLOv5 and our method.
As shown in Table \ref{table1}, our approach proved to be better than the prediction with YOLOv5 trained on synthetic data only.
Since YOLOv5 was trained only on synthetic data, the model could not predict well on real-world images. 
Moreover, since localization and classification of 12 classes were performed simultaneously, the model had no choice but to solve complex problems and was difficult to predict.
On the other hand, our approach consists of two-stage and separates localization and classification so that the model can solve problems easily.
In addition, it was transformed into a photo-realistic image and trained to make predictions in real-world images.
Examples of the predicted class and bounding box using our approach is shown in Fig. \ref{fig-label4}.
\begin{table}[t!]
\centering
\caption{The performance on IEEE BigData 2022 VOD Challenge test set}
\label{table1}
\begin{tabular}{c|ccc} 
\toprule
Method & WmAP(Test 1) & WmAP(Test 2) & Weighted Average Score  \\ 
\hline
YOLOv5 & 0.282       & 0.285       & 0.284                   \\
Ours   & \textbf{0.374}       & \textbf{0.357}       & \textbf{0.365}                   \\
\bottomrule
\end{tabular}
\end{table}

%% file: 06conclusion.tex
\section{Conclusion}
\begin{comment}
우리는 가상의 데이터를 이용하여 실제 환경의 이미지에서 차량의 종류 및 방향을 탐지하는 방법을 제안하였다.
2-stage 모델로 나눠서 학습을 하기 위해 우리는 각 객체에 대한 지역, 라벨, 이미지 정보를 \text{\metatable}로 표현하였고,클래스를 간소화시켜 객체의 지역 정보를 예측하는 방법을 통해 데이터셋의 불균형을 해결하고자 하였다.
더불어, \text{\metatable}의 객체 이미지들을 실제 이미지로 변환하여 모델의 예측력을 높여 실제 환경에서의 이미지에서도 탐지를 잘 할 수 있게 만들었다.

우리는 YOLOv5로 예측했을 때보다 우리의 접근 방식이 더욱 좋다는 것을 증명하였고, 가상의 이미지만으로도 좋은 예측을 이뤄낼 수 있다는 것을 증명하였다.
우리의 주요 한계는 차에 대한 방향을 정확히 판단하는데 부족한 점이 있다는 것이다.
그러나 우리의 접근 방법을 기반으로 다양한 기능 적용을 통해 더 나은 정확도를 향상할 수 있는 통찰력을 제공할 수 있다.
%우리의 방법을 적용하여 Vehicle class and Orientation Detection Challenge 2022에서 4등을 달성하였다.
%우리의 방법을 기반으로 다양한 기법들을 추가하여 합성 이미지만으로도 실제 환경에서 예측을 잘 할 수 있게 되어 훈련 데이터에 대한 의존도가 낮춰지길 희망한다.
%말을 더 추가해야할듯
\end{comment}
We proposed a method to detect the class and orientation of the vehicle in images of a real-world using synthetic data.
To train by dividing into 2-stage model, we stacked each object's region, label, and image information. 
We represented them in a meta-table and tried to solve the dataset imbalance by simplifying the class to predict the region information of the objects.
Furthermore, our model can better predict and detect images in real-world data by converting the object images in the meta-table into photo-realistic images.
We have demonstrated that our approach is better than prediction with vanilla YOLOv5 and that synthetic images alone can make good enough predictions.
The main limitation of our approach is the need for a more accurate prediction of the car's orientation.
However, our approach can provide insights that can improve accuracy with feature adaptions.
%We applied our approach to achieve 4\textsuperscript{th} place in the Vehicle class and Orientation Detection(VOD) Challenge 2022.